\documentclass[letterpaper, 10 pt, conference]{ieeeconf}  

\IEEEoverridecommandlockouts                              

\usepackage[utf8]{inputenc} 
\usepackage[T1]{fontenc}    
\usepackage{hyperref}       
\usepackage{url}            
\usepackage{booktabs}       
\usepackage{amsfonts}       
\usepackage{nicefrac}       
\usepackage{microtype}      
\usepackage{xcolor}         
\usepackage{amsmath}   
\usepackage{amssymb} 
\usepackage{bm}        
\let\labelindent\relax
\usepackage{enumitem}
\usepackage{multirow} 
\usepackage{graphicx} 
\usepackage{array} 
\usepackage{amsmath} 
\usepackage{diagbox}  
\usepackage{makecell} 
\usepackage{caption}
\usepackage{amsmath, amssymb, bm} 
\usepackage{wrapfig}

\title{\LARGE \bf
GAF: Gaussian Action Field as a 4D Representation for Dynamic World Modeling in Robotic Manipulation}

\author{%
    Ying Chai*\textsuperscript{\hspace{0.6em}\rm 1},
    Litao Deng*\textsuperscript{\hspace{0.6em}\rm 2,3},    
    Ruizhi Shao\textsuperscript{\hspace{0.6em}\rm 1},
    Jiajun Zhang\textsuperscript{\hspace{0.6em}\rm 1},
    Kangchen Lv\textsuperscript{\hspace{0.6em}\rm 1},
    Liangjun Xing\textsuperscript{\hspace{0.6em}\rm 1},\\
    Xiang Li$^{\dagger}$\textsuperscript{\hspace{0.6em}\rm 1}
    Hongwen Zhang$^{\dagger}$\textsuperscript{\hspace{0.6em}\rm 2}\textbf{,}
    Yebin Liu$^{\dagger}$\textsuperscript{\hspace{0.6em}\rm 1}\\
  \textsuperscript{\rm 1}Tsinghua University 
  \textsuperscript{\rm 2}Beijing Normal University 
  \textsuperscript{\rm 3}Shadow AI 
  \thanks{*Equal contributions. $^{\dagger}$Corresponding author}
  \thanks{This work was supported by the National Natural Science Foundation of China (NSFC) No.62125107. 
}
}

\begin{document}
\maketitle
\thispagestyle{empty}
\pagestyle{empty}

\begin{abstract}
Accurate scene perception is critical for vision-based robotic manipulation. Existing approaches typically follow either a Vision-to-Action (\textit{\textbf{V-A}}) paradigm, predicting actions directly from visual inputs, or a Vision-to-3D-to-Action (\textit{\textbf{V-3D-A}}) paradigm, leveraging intermediate 3D representations. However, these methods often struggle with action inaccuracies due to the complexity and dynamic nature of manipulation scenes. 
In this paper, we adopt a \textit{\textbf{V-4D-A}} framework that enables direct action reasoning from motion-aware 4D representations via a Gaussian Action Field (GAF). GAF extends 3D Gaussian Splatting (3DGS) by incorporating learnable motion attributes, allowing 4D modeling of dynamic scenes and manipulation actions. To learn time-varying scene geometry and action-aware robot motion, GAF provides three interrelated outputs: reconstruction of the current scene, prediction of future frames, and estimation of init action via Gaussian motion. 
Furthermore, we employ an action-vision-aligned denoising framework, conditioned on a unified representation that combines the init action and the Gaussian perception, both generated by the GAF, to further obtain more precise actions. 
Extensive experiments demonstrate significant improvements, with GAF achieving +11.5385 dB PSNR, +0.3864 SSIM and -0.5574 LPIPS improvements in reconstruction quality, while boosting the average +7.3\% success rate in robotic manipulation tasks over state-of-the-art methods.
\end{abstract}

\section{Introduction}
Effective perception is fundamental to robotic manipulation in unstructured 3D environments.
Recent advances in vision-based methods have enabled robots to infer actions directly from visual observations by leveraging powerful foundation models~\cite{kirillov2023segment, depth_anything_v1, dosovitskiy2021imageworth16x16words}, which facilitates the high-level scene understanding and robotic manipulation.
Existing approaches for vision-based manipulation can be broadly categorized into two paradigms. \textit{\textbf{V-A}} (vision-to-action) paradigm~\cite{chi2023diffusion, shridhar2024generative, Black2023ZeroShotRM} directly map RGB observations to action sequences. While these methods benefit from end-to-end learning, they rely on implicit scene understanding and lack the modeling of 3D world. 
\textit{\textbf{V-3D-A}} (vision-to-3D-to-action) paradigm~\cite{gervet2023act3d, ze20243ddiffusionpolicygeneralizable, ze2023gnfactor} incorporate 3D representations such as point clouds~\cite{gao2024riemann, Chen2023PolarNet3P} and voxel grids~\cite{Shridhar2022PerceiverActorAM,Liu2024VoxActBVA} to enable explicit geometric reasoning. 
By incorporating such structured 3D representations, V-3D-A methods can capture accurate spatial relationships within the scene, leading to higher success rates. Nonetheless, 3D-based methods often require 3D datasets which are much less available than image datasets. 
Despite their differences, both paradigms share a common limitation: they fail to capture the temporal evolution of scene geometry, leading to an inherent mismatch between static scene understanding and dynamic action generation.

\begin{figure}[t!]
    \raggedright
    \includegraphics[width=\columnwidth]{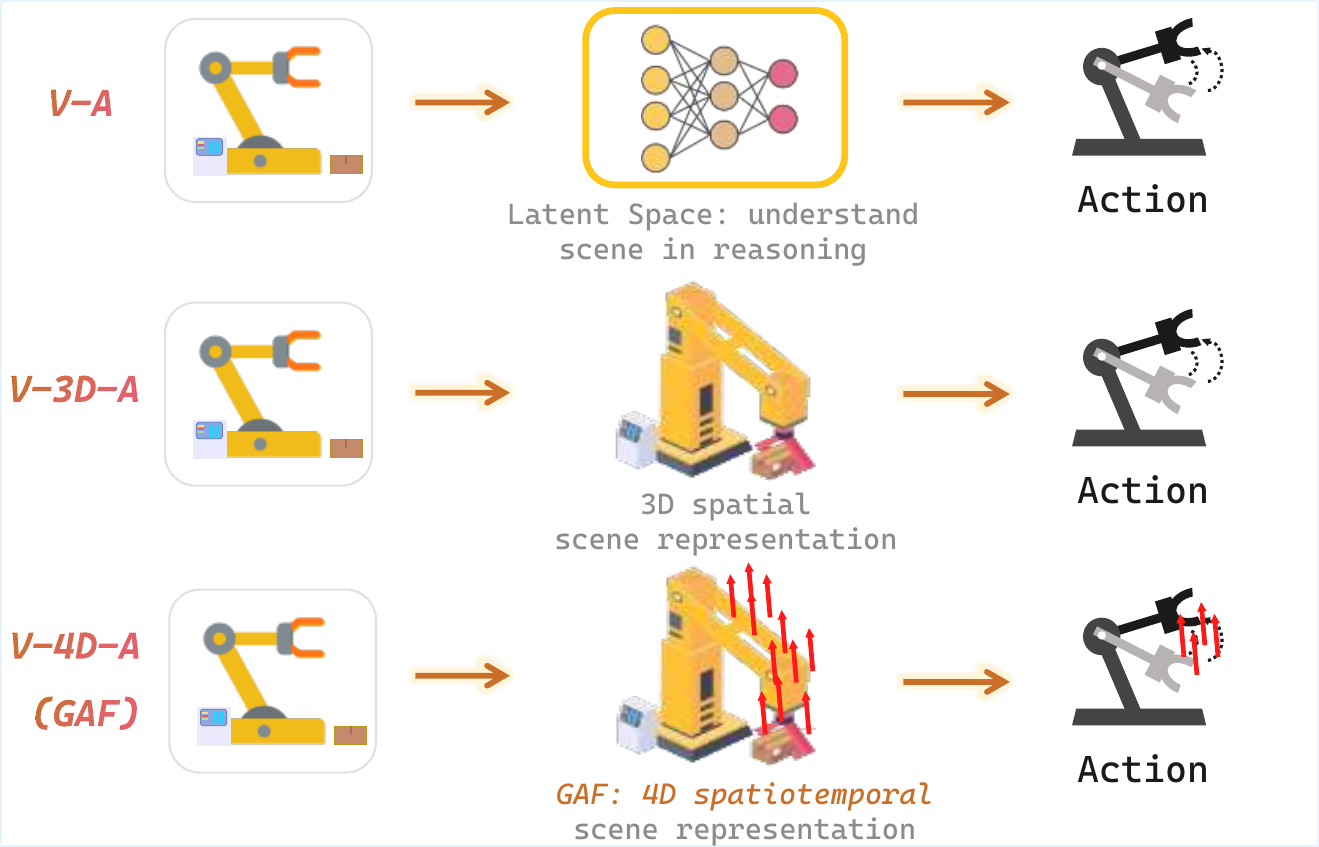}
    \caption{Comparisons between the previous \textit{\textbf{V-A}} paradigm, \textbf{\textit{V-3D-A}} paradigm and our proposed \textit{\textbf{V-4D-A}} method GAF.}
    \label{fig: intro}
\end{figure}

In response to this challenge, the \textit{\textbf{V-4D-A}} (vision-to-4D-to-action) paradigm has recently emerged, which augments 3D representations with motion information to capture the temporal evolution of scenes, as illustrated in Fig.~\ref{fig: intro}.
Unlike static representations that passively encode geometry, these methods aim to guide robotic action planning by modeling how scene geometry, including the robot itself, may evolve over time.
Such 4D dynamic perception enable more intuitive action inference since the scene motion inherently contains the movement trend information. 

Building on this paradigm, recent efforts~\cite{zhang2025vla,niu2025pre} have begun to explore 4D representations for robotic manipulation tasks.
Most of these approaches model 4D representations primarily by leveraging the sequential nature of video frames, without explicitly embedding dynamic information into the scene representation.
Some recent works address this limitation by integrating dynamic information more naturally into the scene perception through Gaussian-based world model: ManiGaussian \cite{lu2024manigaussian} and GWM~\cite{lu2025gaussianworldmodel} deform current Gaussian for future scene consistency to further supervise action prediction. However, their methods only use the future states as volumetric priors to guide policy training. Their implicit use of Gaussian is less effective due to the low fidelity. 
In this end, we seek to construct a more explicit 4D representation by directly modeling the temporal transformation between high-fidelity current and future states, which enables reliable action inference grounded in accurate dynamic scene understanding.

In this paper, we introduce the \textbf{Gaussian Action Field (GAF)} as a concrete implementation of the V-4D-A paradigm. 
GAF is built on 3D Gaussian Splatting (3DGS)~\cite{kerbl20233dgaussiansplattingrealtime} due to its strong geometric fidelity and its differentiable rendering mechanism, which allows supervision from RGB video frames without requiring ground-truth 3D data.
To extend 3DGS to dynamic scenes, GAF introduces a learnable motion attribute, which encodes the temporal displacement of each Gaussian point to capture evolving scene geometry and infer motion within a unified representation.
GAF produces three interrelated outputs, each corresponding to a key stage in the perception-to-action pipeline:
The current Gaussian provides a view-consistent encoding of the present scene. 
The future Gaussian predicts how the scene evolves by applying the learned motion attributes. 
The init action is computed through point cloud registration between the current and future Gaussian. 
These components together allow GAF to connect dynamic visual perception with action generation, forming a complete V-4D-A framework. 
To further improve action quality, we introduce an action-vision-aligned denoising module that refines init actions using action-aware visual guidance produced by GAF. 
This overall approach resonates with the concept of the world model~\cite{ha2018recurrent}, modeling future scene dynamics to support downstream decision-making.

GAF operates in a fully feed-forward manner and supports real-time execution on a single GPU during manipulation. 
Extensive experiments demonstrate that our method enables high-quality scene reconstruction, accurate robotic manipulation and spatial generalization, outperforming V-A and V-3D-A baselines. Moreover, the approach has been successfully deployed in real-world environments, demonstrating its practical feasibility. Contributions are summarized as follows:

\begin{itemize}
    \item We introduce a V-4D-A method GAF, extended 3DGS with motion inference, unifying dynamic scene evolution and future-oriented action prediction.
    \item We propose an action-vision-aligned denoising framework to enable action refinement within a unified image space, enhancing the accuracy of action prediction.
    \item We validate our method on robotic manipulation tasks, where it achieves state-of-the-art performance in both scene reconstruction quality and action generation.
\end{itemize}

\section{Related Work}
\subsection{Vision-based Robot Learning}
Vision~\cite{Black2023ZeroShotRM,gao2024riemann,Chen2023PolarNet3P,Chen2024G3FlowG3} plays a pivotal role in enabling robots to perceive and interact with their environments. 
2D methods that rely on image inputs are the earliest to emerge and are supported by the most extensive datasets. Several of these methods have demonstrated remarkable performance. The Google RT series~\cite{brohan2023rt2visionlanguageactionmodelstransfer},  IGOR~\cite{chen2024igor} and ViLBERT~\cite{lu2019vilbertpretrainingtaskagnosticvisiolinguistic} have achieved impressive results through massive-scale data training, while Diffusion Policy~\cite{chi2023diffusion} and GENIMA~\cite{shridhar2024generative} utilizes diffusion model to generate action sequence. 
These methods often struggle with accurately capturing precise 3D spatial relationships, limiting their effectiveness in high-precision tasks~\cite{kloss2020accurate}. 
In contrast, 3D methods explicitly model volumetric geometric structures: Act3D~\cite{gervet2023act3d} and RVT series~\cite{goyal2024rvt2learningprecisemanipulation} utilize point clouds for scene representation. Others, including PerAct~\cite{Shridhar2022PerceiverActorAM}, VoxAct~\cite{Liu2024VoxActBVA} and GNFactor~\cite{ze2023gnfactor} adopt voxel grids to encode the scene geometry.These 3D representations enabling accurate spatial structure and higher success rate.

These methods neglect the fact that, in addition to complex geometric structures and spatial relationships, robot learning also requires the consideration of time as a crucial dimension. 
Unlike previous methods, our approach explicitly models temporal dynamics, allowing accurate 3D scene understanding and action prediction solely from RGB video frames.

\subsection{World Model}

World models obtain environmental knowledge by constructing an internal representation that simulates how the world evolves~\cite{ha2018recurrent,gao2024enhance,hafner2019dream,hafner2023mastering}. 
These methods successfully encode scene dynamics~\cite{hansen2023td,seo2023masked} by predicting future states from current observations.
Previous approaches utilize auto-encoding to learn a latent space for predicting future states. The implicit nature of their feature representations, combined with high data requirements, limits the effectiveness and practical applicability of these methods.
Recent methods enhance generalization by using explicit representations in image~\cite{mendonca2023structured} and language~\cite{lu2023thinkbot,Zhen20243DVLAA3}: ManiGaussian~\cite{lu2024manigaussian} and GWM~\cite{lu2025gaussianworldmodel} adopt Gaussian as state representation and leverage predicted future Gaussian to guide policy training.

Unlike action-conditioned world models~\cite{lu2024manigaussian,lu2025gaussianworldmodel} that use future prediction as auxiliary supervision, our GAF is vision-driven and task-oriented: it forecasts scene evolution directly from visual inputs and explicitly infers actions from the predicted future geometry, enabling direct action extraction and more efficient inference.


\section{Method}
\label{methods}
In this section, we introduce GAF, its implementation, and its application in robotic manipulation tasks.
Sec.~\ref{sec:definition} defines GAF representation and its outputs. 
Sec.~\ref{sec:implementation} details GAF network design. 
Sec.~\ref{sec:manipulation} illustrates how the outputs of GAF are used to generate executable actions for a complete robotic manipulation task.

Our overall pipeline consists of two stages: 4D GAF scene reconstruction and action refinement via diffusion. In the first stage, GAF reconstructs dynamic scenes using only sparse multi-view RGB images and their corresponding intrinsics, without requiring camera extrinsics. In the second stage, the refinement module leverages both the reconstructed Gaussians and the known camera extrinsics (obtained via calibration) to render action-conditioned visual guidance for denoising.

\subsection{GAF Representation}
\label{sec:definition}
We define the Gaussian Action Field (GAF) as a unified spatiotemporal representation that associates each Gaussian primitive \(\mathbf{g(x)}\) at time step \(t\) with both geometric attributes and motion dynamics. Formally, GAF is parameterized by a continuous function \(\mathcal{F}_{\Theta}\):
\begin{equation}
\mathcal{F}_{\Theta}: \left\{{g(x)}, t\right\} \mapsto \left\{\mu, \Delta\mu, f\right\},
\end{equation}
where \(\mu \in \mathbb{R}^3\) denotes the 3D position of the Gaussian, \(\Delta\mu \in \mathbb{R}^3\) is displacement vector indicating temporal motion, and appearance parameters \(f = \{c, \sigma, r, s\}\) represents the color, opacity, rotation, and scale attributes of each Gaussian.

GAF produces three types of outputs: the current Gaussian, the future Gaussian, and the init action. These outputs are generated from different combinations of parameters. The current Gaussian is constructed from the position and appearance parameters $\left\{\mu, f\right\}$, representing the present scene. The future Gaussian is obtained by applying the predicted motion to the position parameters and combining the result with the same appearance features, yielding $\left\{\mu+\Delta\mu, f\right\}$. The init action is estimated through point cloud matching between the current and future manipulation-related Gaussian.

\subsection{GAF Architecture}
\label{sec:implementation}
\begin{figure}[t!]
    \centering
    \includegraphics[width=\columnwidth]{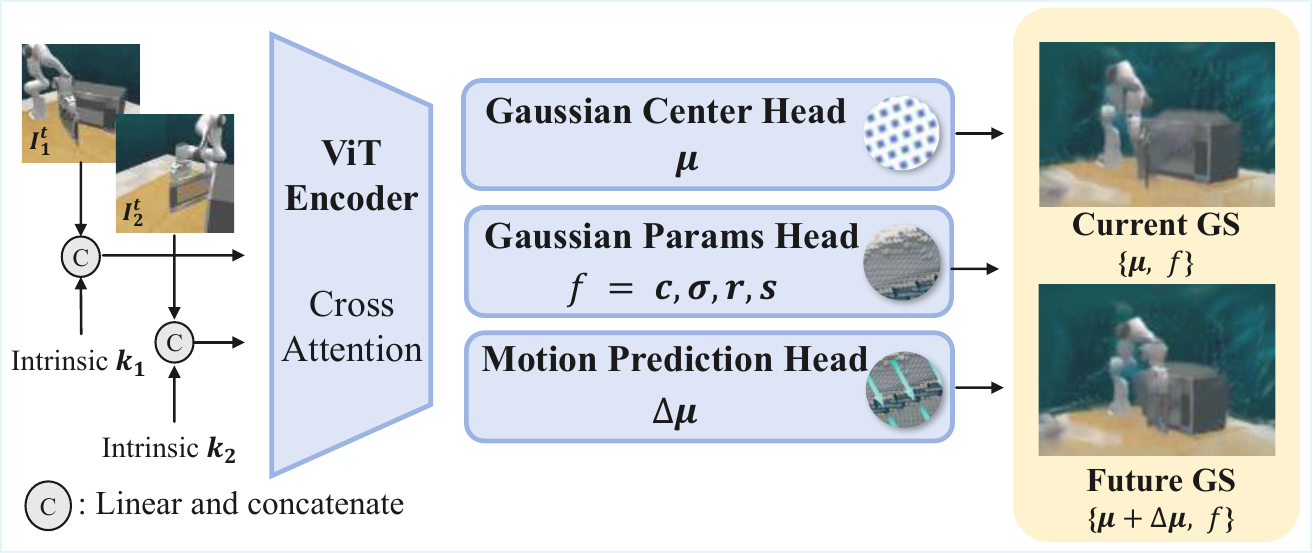}
    \caption{\textbf{Overview of GAF reconstruction.}
    Given sparse multi-view images, a Vision Transformer extracts hybrid scene features, which are decoded by three heads to predict Gaussian positions, motions, and appearance parameters. 
    }
    \label{fig: GAF}
\end{figure}

The Gaussian Action Field (GAF) architecture unifies scene representation, scene dynamics, and action reasoning. Our goal is to reconstruct motion-augmented Gaussian directly from sparse, unposed RGB inputs, enabling downstream manipulation control. Fig.~\ref{fig: GAF} illustrates the overall design.

\begin{figure*}[t!]
    \centering
    \includegraphics[width=\textwidth]{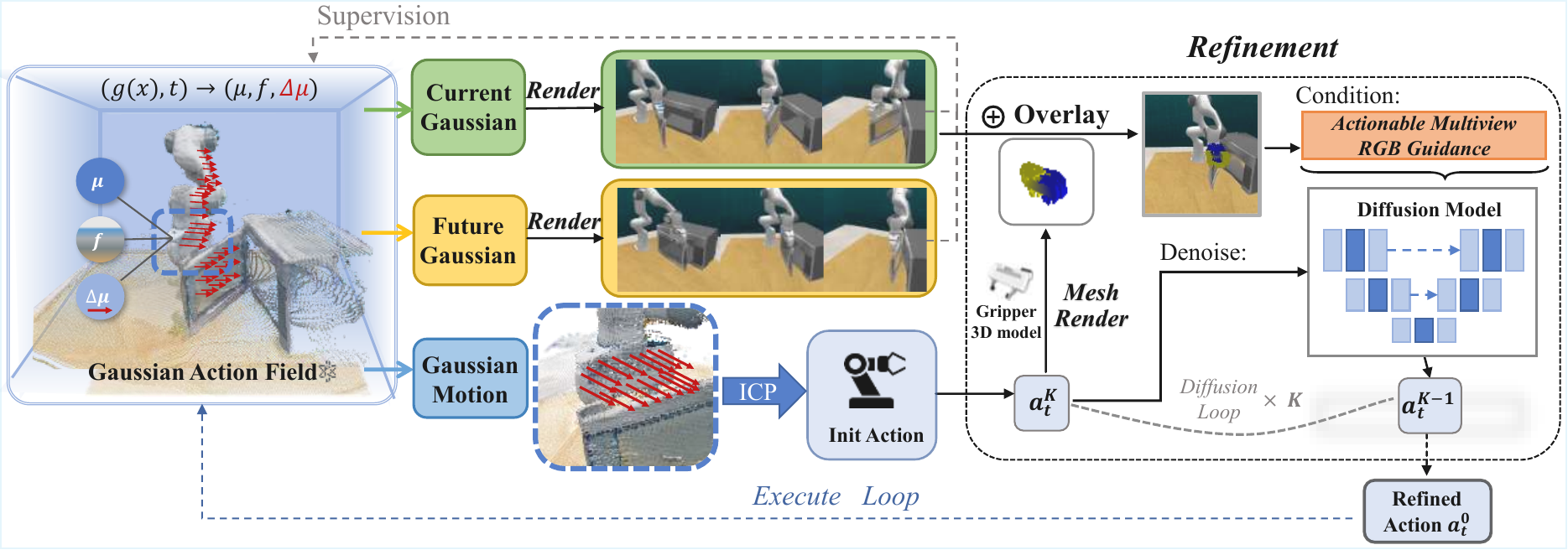}
    \caption{\textbf{Manipulation pipeline.}
    GAF outputs are then used as conditions for a action-vision-aligned denoising framework to generate executable motion. The process repeats iteratively until the task completes.
    }
    \label{fig: mani}
\end{figure*}

\paragraph{Dynamic Gaussian Reconstruction}
GAF adopts a geometry-agnostic, pose-free approach for dynamic scene reconstruction, in contrast to traditional methods such as NeRF~\cite{mildenhall2020nerf} and 3DGS~\cite{kerbl20233dgaussiansplattingrealtime}, which rely on dense camera poses or strong geometric priors like cost volumes, epipolar constraints. 
Our architecture directly reconstructs high-fidelity motion-augmented Gaussian by two input views in a canonical space aligned with the first input view. This is achieved using a feed-forward network that includes a vision transformer backbone and three specialized heads.

Specifically, given two unposed \(H \times W\) images and their corresponding intrinsics \(\{I_v^t,k_v^t\}_{v=1}^V\) at timestep $t$, we tokenize images into patch sequences and concatenate them. The resulting tokens are fed into a shared-weight Vision Transformer with cross-view attention to extract features.

For scene representation, we employ a decoupled two-head design $\mathcal{H}_{\text{Gauss}}=\{h_{\text{center}},h_{\text{feature}}\}$ based on the DPT architecture\cite{Ranftl2021} to process the features: the Gaussian Center Head $h_{\text{center}}$ predicts only Gaussian point positions, the Gaussian Param Head $h_{\text{feature}}$ estimates the remaining Gaussian parameters named as appearance parameters \(f = \{c, \sigma, r, s\}\). The process can be formulated as:
\begin{equation}  
\mathcal{H}_{\text{Gauss}}\left(\text{ViT}(\{\bm{I}_v^t, \bm{k}_v^t\})\right)_{v=1}^V = \{{\mu}_j^{t}, c_j^{t}, \sigma_j^{t}, r_j^{t},s_j^{t}\}_{j=1}^{V \times H \times W},  
\label{eq:gauss_head}  
\end{equation}  
For scene dynamics, we introduce a Motion Prediction Head $h_{\text{motion}}$ following the same DPT-based architecture\cite{Ranftl2021} as Gaussian Center Head. $h_{\text{motion}}$ predicts the per-point displacement \(\Delta\mu_j^{t \rightarrow t+\Delta t}\), representing the motion of each Gaussian over a future interval $\Delta t$:
\begin{equation}  
h_{\text{motion}}\left(\text{ViT}(\{\bm{I}_v^t, \bm{k}_v^t\})\right)_{v=1}^V = \{\bm{\Delta\mu}_j^{t \rightarrow t+\Delta t}\}. 
\label{eq:disp_head}  
\end{equation}  
The predicted displacement \(\Delta\mu_j^{t \rightarrow t+\Delta t}\) are added to \(\mu_j^t \) to obtain the future Gaussian positions \(\mu_j^{t+\Delta t } \). These displaced centers are fused with the appearance parameters \((c_j^{t}, \sigma_j^{t}, r_j^{t},s_j^{t})\) to form the future Gaussian. 

Deriving the current Gaussian and future Gaussian , we can render $M$ novel view images for the current state $\{\hat I_v^t\}_{v=1}^M$ and future state $\{\hat I_v^{t+\Delta t}\}_{v=1}^M$ using alpha-blend rendering, $M$ denote the number of synthesized views. To be specific, the pixel color at location \(\mathbf{p}\) is computed by:
\begin{equation}
C(\mathbf{p}) = \sum_{i=1}^N \alpha_i c_i \prod_{j=1}^{i-1} (1 - \alpha_j),
\quad
\label{eq:alpha-blend}
\end{equation}
where \(C\) is the rendered image, \(N\) denotes the number of Gaussian, \(\alpha_i = \sigma_i e^{-\frac{1}{2} (\mathbf{p} - \mu_i^{2d})^\top \Sigma_i^{-1} (\mathbf{p} - \mu_i^{2d})}\) represents the 2D density in the splatting process, and \(\Sigma_i\) stands for the covariance matrix acquired from the rotation \(r\) and scales \(s\). $M$ denote the number of synthesized views.

This allows for direct RGB video frames supervision for the entire Dynamic Gaussian Reconstruction. 
We learn \(\mathcal{F}_{\Theta}\) by minimising the following~\cite{ye2024noposplat}:
\begin{equation}  
\mathcal{L}_{\text{GAF}} = \mathcal{L}_{\text{LPIPS}}^t+\mathcal{L}_{\text{MSE}}^t+\mathcal{L}_{\text{LPIPS}}^{t+\Delta t}+\mathcal{L}_{\text{MSE}}^{t+\Delta t}. 
\end{equation}  
where $\mathcal{L}^t$ enforces geometric fidelity to current observations and $\mathcal{L}^{t + \Delta t}$ regularizes future state prediction. They are aggregated into a unified objective, facilitating the joint optimization of motion-augmented Gaussian reconstruction.

With access to 4D Gaussian-based perception at both the current and future time steps, we are able to extract more concrete actions from such representations.

\paragraph{Init Action Computation}
Since our task centers on manipulation, our attention is directed toward the motion of the robotic arm itself, especially the end-effector. Due to its rigid nature, we extract the gripper-part Gaussians \(\bm{\mu}_{\text{gripper}}\) and future state Gaussians 
\((\bm{\mu+\Delta \mu})_{\text{gripper}}\), and estimate a rigid transformation \(T^{t \rightarrow t+\Delta t} \in \text{SE}(3)\)  using ICP~\cite{Segal2009GeneralizedICP}:
\begin{equation}
T^{t \rightarrow t+\Delta t} = \arg\min \sum_{k \in gripper} \lVert T(\mu_k) - (\bm{\mu+\Delta \mu})_{k} \rVert^2
\end{equation}
$T^{t \rightarrow t+\Delta t}$ denotes the end-effector transformation matrix over a time interval $\Delta t$, providing an explicit representation of scene dynamics. We interpolate it into a sequence of transformations across discrete timesteps. This sequence forms the init action $a_{init}$, describing the transition from the current frame $T$ to the future frame $t+\Delta t$.

\subsection{Manipulation with GAF}
\label{sec:manipulation}

After introducing the definition and architecture of GAF, we now describe how it is deployed in manipulation tasks. 
As the GAF module is trained solely on visual data without real action supervision, the predicted init actions are only visually plausible and often lack physical feasibility during interaction. To address this, we adopt a denoising network and incorporate a small amount of real action data to refine the initial predictions and ensure physically consistent execution. 

As illustrated in Fig.~\ref{fig: mani}, to fully exploit GAF's outputs, we adopt action-vision-aligned denoising framework inspired by R\&D~\cite{vosylius2024render}. Specifically, we use a rendering process to visualize the spatial consequences of candidate actions. 
To determine the pose in which the gripper should be rendered, we first compute its pose in the world frame as $T_{g_{new}}^w = T_{g}^w \times a$, where $a$ is a relative transformation representing the intended action. Then, by utilising the camera $c$'s extrinsic matrix $T_{w2c}$ and intrinsic matrix $K$, we can reposition the gripper's CAD model in the camera's frame and render an image of it, creating the rendered action representation $R^c$:

\begin{equation}
    R^c = Render(T_{w2c} \times T_{g}^w \times a, K^c)
    \label{eq:rendered_actions}
\end{equation}

For each denoising step of duration \(\Delta t\) , we project the gripper positions resulting from the init action $a_{init}$ to pixel coordinates and render gripper mesh onto current multi-view RGB images $\{\hat I_v^t\}_{v=1}^M$.
These multiview rendered action representations $R^c$ create a unified representation, termed \textit{Actionable Multiview RGB Guidance}, which integrates the visual 3D observations with the temporally predicted actions. 
Such visual cues guide the diffusion model to minimize the following constraints:\textit{}
\begin{equation}
\mathcal{L}_{refine} = L1(D, D^{gt}) + L1(\epsilon, \epsilon^{gt}) + BCE(g, g^{gt})
\label{eq:diffusion}
\end{equation}
where $D$ represents denoising direction of gripper. $\epsilon$ is the noise added to the end-effector action. $g$ is a binary variable that represents gripper's opening-closing action. $D^{gt},\epsilon^{gt},g^{gt}$ are their ground truth labels respectively. 
The denoised action sequence represents target end-effector poses in the world frame. These poses can be executed via inverse kinematics to reach the desired positions. Upon execution, the environment is updated and new observations are collected, enabling the next iteration of the control loop.

The entire control loop, comprising GAF's 4D scene representation, denoising framework and execution, is repeated iteratively until the manipulation task is completed. This closed-loop framework enables continuous adaptation to dynamic scene changes, leveraging GAF’s spatiotemporal reasoning to maintain robust performance under occlusion and interaction uncertainties.

\section{Experiments}
To validate the effectiveness of our approach, we perform extensive evaluations in both simulation and real-world environments. Furthermore, we conduct ablation studies, generalization tests, and multi-task experiments to thoroughly demonstrate the model's capabilities.

\subsection{Comparison on Simulated Environments}

\textbf{Experiment Setup.}
We evaluate our method on RLBench\cite{james2020rlbench} across 9 tasks, covering manipulation challenges including fine-grained placement, occlusion-rich interactions and articulated object handling. To ensure generalization, we randomly initialize the objects in the environment and collect 20 demonstrations of each individual task and tests across 100 different unseen poses for each task. 

\textbf{Baselines. }
For baseline selection, we compare our proposed \textbf{\textit{V-4D-A}} paradigm against representative methods from the V-A and V-3D-A categories to highlight its advantages.

In the \textbf{\textit{V-A}} category, we select Diffusion Policy (DP)~\cite{chi2023diffusion} as a representative baseline, given its widespread adoption, open-source availability, and established performance in predicting actions directly from 2D visual inputs.
In the \textbf{\textit{V-3D-A}} category, we use Act3D~\cite{gervet2023act3d} as the baseline. It utilizes RGB-D inputs and leverages depth information along with camera parameters to lift 2D data into 3D space, generating a 3D scene feature cloud for action prediction.

We also include a comparison with ManiGaussian~\cite{lu2024manigaussian}, which, like our method, is based on a Gaussian world model. 
Hyper-parameters such as prediction horizon, observation history or the number of trainable parameters were adjusted to match our method for fair comparison.

\begin{table}[t]
\centering
\begin{minipage}{\columnwidth}
\caption{Success rates (\%) comparison on RLBench.}
\centering
\setlength{\tabcolsep}{1.0mm}
\resizebox{0.9\columnwidth}{!}{%
\begin{tabular}{l|ccccc}
\toprule
\textbf{Method} & \makecell{Toilet\\ Seat Down} & \makecell{Open\\ Grill} & \makecell{Close\\ Grill} & \makecell{Close\\ Fridge} & \makecell{Phone\\ On Base} \\
\midrule
DP~\cite{chi2023diffusion} & 39 & 20 & 36 & 16 & 23 \\
Act3D~\cite{gervet2023act3d} & 60 & 22 & 41 & \textbf{47} & 26 \\
ManiGaussian~\cite{lu2024manigaussian} & 34 & 24 & 38 & 41 & 30 \\
\midrule
\textbf{Ours w/o GAF} & 57 & 16 & 49 & 27 & 31 \\
\textbf{Ours with GAF} & \textbf{71} & \textbf{26} & \textbf{55} & 42 & \textbf{35} \\
\bottomrule
\end{tabular}%
}
\label{tab:success_rate_part1}
\end{minipage}

\vspace{1em}
\begin{minipage}{\columnwidth}
\centering
\setlength{\tabcolsep}{1.0mm}
\resizebox{0.9\columnwidth}{!}{%
\begin{tabular}{l|cccc|c}
\toprule
\textbf{Method} & \makecell{Lift\\ Lid Up} & \makecell{Close\\ Microwave} & \makecell{Push\\ Button} & \makecell{Close\\ Laptop} & \textbf{Avg.} \\
\midrule
DP~\cite{chi2023diffusion} & 81 & 62 & 61 & 64 & 44.7 \\
Act3D~\cite{gervet2023act3d} & 91 & 73 & 60 & 58 & 53.1 \\
ManiGaussian~\cite{lu2024manigaussian} & 97 & 67 & \textbf{63} & 57 & 50.1 \\
\midrule
\textbf{Ours w/o GAF} & 94 & 79 & 58 & 51 & 51.3 \\
\textbf{Ours with GAF} & \textbf{100} & \textbf{85} & 61 & \textbf{69} & \textbf{60.4} \\
\bottomrule
\end{tabular}%
}
\label{tab:success_rate_part2}
\end{minipage}
\end{table}

\textbf{Results \& Discussion. }
The experimental results in Table~\ref{tab:success_rate_part1} demonstrate that our method outperforms all selected baselines.
Compared to the V-A method Diffusion Policy (DP), our approach achieves a 15.7\% improvement in success rate, particularly in tasks with heavy occlusion like "Toilet Seat Down", which highlights the critical role of 3D perception in robotic manipulation.
Compared to the V-3D-A method Act3D, our method achieves a 7.3\% improvement, indicating the effectiveness of incorporating future-aware 4D dynamic scene representations in manipulation tasks.
Furthermore, when compared to ManiGaussian, our method yields a 10.3\% increase in success rate, further validating the advantages of our explicit action generation and denoising strategy.

\subsection{Comparison on Scene Reconstruction and Prediction}
\label{reconstruction compare}
For scene reconstruction quality, we compare GAF against ManiGaussian~\cite{lu2024manigaussian}, both of which generate Gaussian point clouds for current frame reconstruction and future frame prediction. 
We do not include dynamic reconstruction methods such as 4DGS~\cite{yang2023gs4d} as baselines in this comparison, since their outputs are typically interpolations of the input images across both viewpoint and time. In other words, these methods can only reconstruct Gaussian corresponding to intermediate time points between input frames, rather than predicting future scene states from a given time point.

\begin{figure}[t!]
    \centering
    \includegraphics[width=\columnwidth]{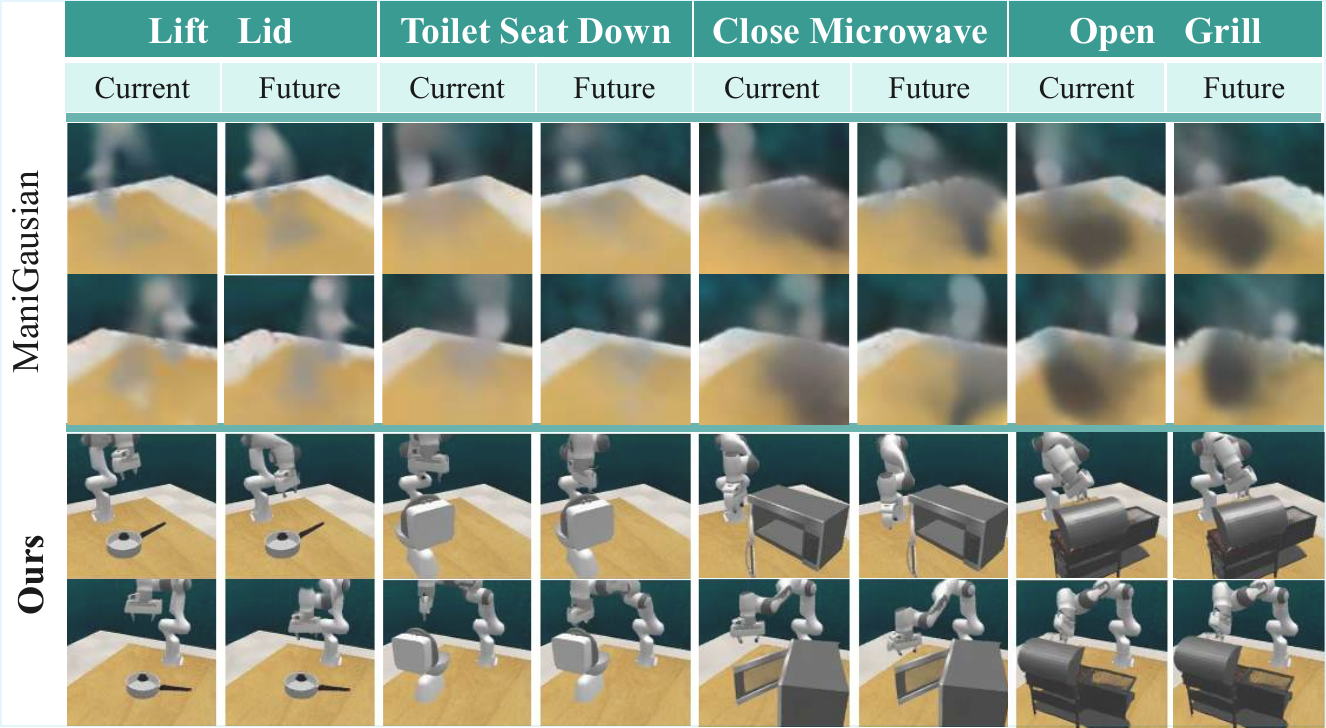}
    \caption{
    Comparison of current scene reconstruction and future scene prediction from novel views.}
    \label{fig: compare_manigaussian}
\end{figure}

\textbf{Qualitative Analysis. }
As shown in Fig.~\ref{fig: compare_manigaussian}, our method achieves superior reconstruction fidelity and novel-view synthesis. ManiGaussian’s renders (up) exhibit blurred textures and incomplete geometric details resulting in ambiguous spatial relationships. In contrast, our renders (down) preserve fine geometric structures, such as the gripper’s articulated joints and object surfaces, even under partial observations. 
This clarity in reconstructing the Gaussian point cloud allows for the extraction of precise end-effector point clouds to calculate the action, which contributes to the fundamental difference compared to ManiGaussian.

\textbf{Quantitative Metrics. }
We further evaluate reconstruction quality using standard metrics: PSNR (photometric fidelity), SSIM (structural similarity), and LPIPS (perceptual consistency). As shown in Table~\ref{tab:reconstruction}, our method outperforms ManiGaussian by +11.5385 dB PSNR, +0.3864 SSIM, -0.5574 LPIPS on average across tasks in current scene reconstruction, and +10.5311 dB PSNR, +0.3856 SSIM, -0.5757 LPIPS in future state prediction. These metrics confirm that our dynamic rendering framework ensures high quality geometric accuracy and temporal coherence.
\begin{table*}[t]
    \caption{Current \& Future Novel view synthesis performance Comparison.}
    \centering
    \renewcommand{\arraystretch}{1.2} 
    \setlength{\tabcolsep}{0.75mm}
    \resizebox{\linewidth}{!}{
    \begin{tabular}{l|ccc|ccc|ccc|ccc}
        \toprule
        \multirow{2}{*}{\textbf{Method}} & 
        \multicolumn{3}{c|}{Close Microwave} & 
        \multicolumn{3}{c|}{Toilet Seat Down} & 
        \multicolumn{3}{c|}{Lift Lid Up} & 
        \multicolumn{3}{c}{\textbf{Average}} \\
        \cmidrule(lr){2-4} \cmidrule(lr){5-7} \cmidrule(lr){8-10} \cmidrule(lr){11-13}
        & PSNR$\uparrow$ & SSIM$\uparrow$ & LPIPS$\downarrow$ & 
        PSNR$\uparrow$ & SSIM$\uparrow$ & LPIPS$\downarrow$ & 
        PSNR$\uparrow$ & SSIM$\uparrow$ & LPIPS$\downarrow$ & 
        PSNR$\uparrow$ & SSIM$\uparrow$ & LPIPS$\downarrow$ \\
        \midrule
        ManiGaussian\cite{lu2024manigaussian} /Now& 16.4274 & 0.3753 & 0.7628 & 16.5628 & 0.3976 & 0.6806 & 16.1492 & 0.4139 & 0.6217 & 16.3798 & 0.3956 & 0.6884 \\ 
        ManiGaussian\cite{lu2024manigaussian} /Future& 16.1368 & 0.3565 & 0.7896 & 15.7953 & 0.3687 & 0.7161 & 15.3727 & 0.3969 & 0.6572 & 15.7683 & 0.3740 & 0.7210 \\ 
        \textbf{Ours / Now} & \textbf{27.0986} & \textbf{0.7976} & \textbf{0.1291} & \textbf{28.1652} & \textbf{0.7779} & \textbf{0.1352} & \textbf{28.4912} & \textbf{0.7705} & \textbf{0.1286} & \textbf{27.9183} & \textbf{0.7820} & \textbf{0.1310} \\ 
        \textbf{Ours / Future} & \textbf{24.5881} & \textbf{0.7650} & \textbf{0.1489} & \textbf{27.2951} & \textbf{0.7655} & \textbf{0.1456} & \textbf{27.0150} & \textbf{0.7483} & \textbf{0.1413} & \textbf{26.2994} & \textbf{0.7596} & \textbf{0.1453} \\ 
        \bottomrule
    \end{tabular}}
    \\[0.75ex]
    {\raggedright \footnotesize $\uparrow$: Higher is better; $\downarrow$: Lower is better.}
    \label{tab:reconstruction}
\end{table*}

\subsection{Ablation Study}
\label{sec:ablation}

\begin{figure*}[t!]
    \centering
    \includegraphics[width=0.8\textwidth]{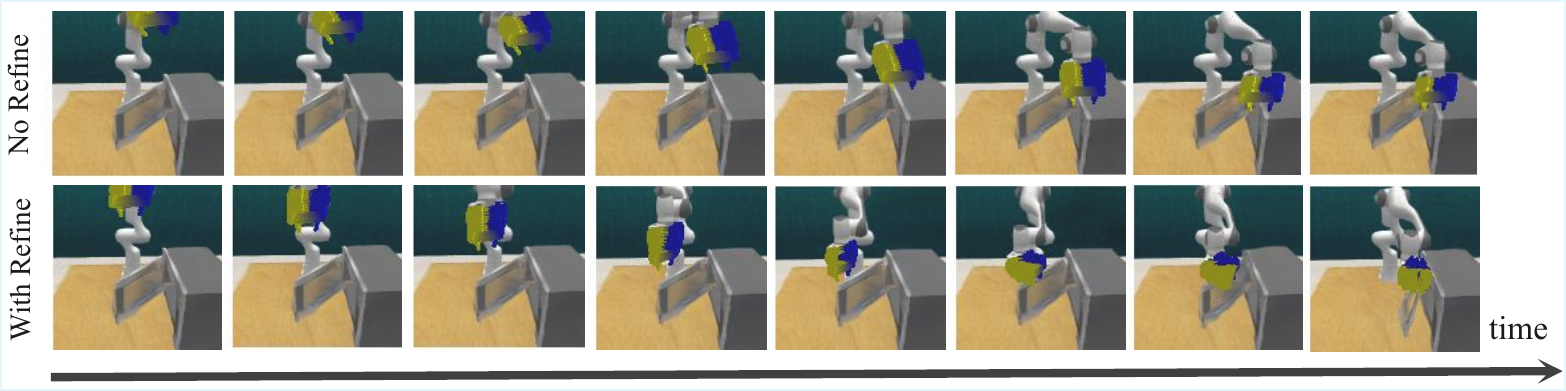}
    \caption{\textbf{Ablation on Denoising Framework} The upper image shows a failed experiment without action denoising, while the lower image depicts a successful experiment after action denoising.}
    \label{fig: ablation_action}
\end{figure*}

\textbf{Ablation on GAF. }
To evaluate the contribution of GAF, we remove this component and directly predict actions from input images using a diffusion model that denoises from random noise, without multi-view rendering or init action priors.
With the ablation study results demonstrate in Table~\ref{tab:success_rate_part1}, W/o GAF method drop our full method by -9.1\% in average success rate across tasks. These results demonstrate that GAF's 4D representation improves action prediction accuracy.

\textbf{Ablation on Denoising Framework. }
We set up a control group that directly executes the init actions.
Experimental observations in Fig.~\ref{fig: ablation_action} show that the arm fails to reach the correct position but continues pushing forward without action refinement.
These results demonstrate that although GAF performs well in the early stages, reconstruction errors caused by partial occlusions lead to misaligned contacts during interaction, ultimately resulting in task failure.
Therefore, action refinement is crucial for the successful completion of the entire task.

\subsection{Spatial Generalization}
\label{sec: spa}
\begin{figure}[t]
    \centering
    \includegraphics[width=\columnwidth]{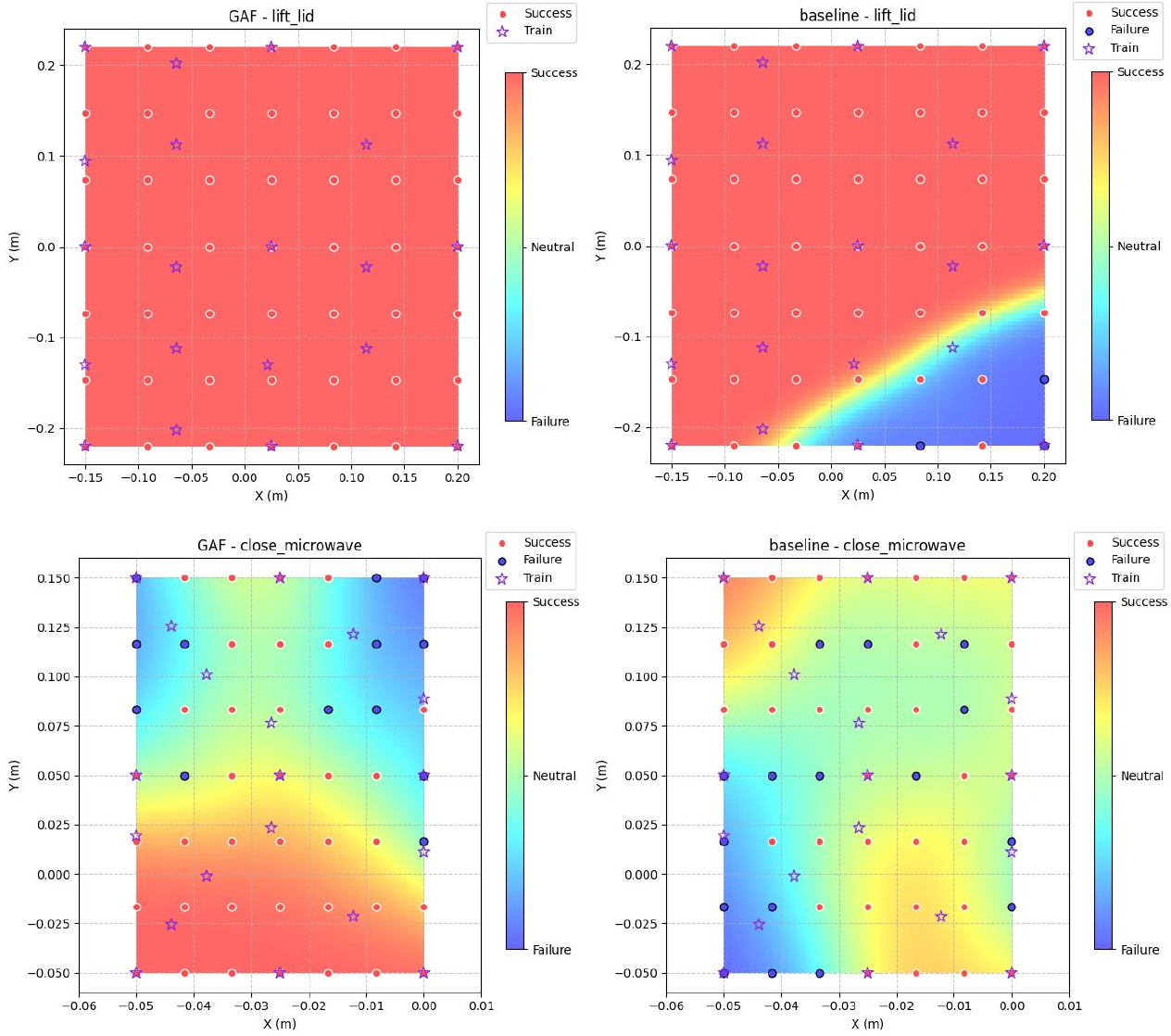}
    \caption{\textbf{Spatial Generalization.} Outcome of our method and baseline trained on 20 demonstrations (purple stars). Red and blue colors representing successes and failures, respectively.}
    \label{fig: spa}
\end{figure}

\textbf{Experiment Setup. }
To evaluate whether the model can achieve  spatial generalization to object positions, we adopt a data collection strategy to ensure comprehensive spatial coverage of the operational workspace on $2$ tasks on RLBench. 
During evaluation, we employ a grid sampling methodology across the entire workspace, which guarantees sufficient spatial variation and measurement consistency.

\textbf{Results \& Discussion. }
As illustrated in Fig.~\ref{fig: spa}, the baseline DP encounters challenges when objects are placed along the boundaries and corners of the workspace. In contrast, our method achieves superior spatial generalization capability even when objects are placed on boundaries. Besides, our method is less sensitive to corners.

\subsection{Multi-task Evaluation}
\begin{table}[t]
\caption{Success rates (\%) with  deltas of our method and baseline trained jointly on 4 tasks on RLBench.}
\centering
\setlength{\tabcolsep}{1.0mm}
\resizebox{\columnwidth}{!}{%
\begin{tabular}{l|c|c|c|c|c}
\toprule
\textbf{Method} & 
\makecell{Toilet\\Seat Down} & 
\makecell{Close\\Microwave} & 
\makecell{LIFT\\LID Up} & 
\makecell{Close\\Laptop} & 
\textbf{Average} \\ 
\midrule
DP~\cite{chi2023diffusion} & 55 (+16) & 50 (-12)  & 39 (-42)  & 18 (-46) & 40.5 (-21)\\
\midrule
Ours & \textbf{59 (-12)} & \textbf{85 (+0)} & \textbf{57 (-43)} & \textbf{79 (+10)} & \textbf{70 (-11.25)} \\
\bottomrule
\end{tabular}%
}
\label{tab:success_rate}
\end{table}

\textbf{Experiment Setup. }
In previous experiments, we trained the model for each task. To validate the generalization capabilities, we test its capacity to learn multiple tasks simultaneously. 
We train a single network using data collected from 4 RLBench tasks, 20 demonstrations each. Object positions are randomly initialized in both data collection and model evaluation phase.

\textbf{Results \& Discussion. }
As table~\ref{tab:success_rate} illustrated (The values in parentheses represent the performance change compared to single-task training), our method's average success rate only declines 11.25\%. Our success rate exhibits the most significant decline in the "lift lid up" task, which is markedly distinct from the other three tasks. Nevertheless, in comparison to the substantial 21\% decline observed in the baseline, our method demonstrates considerably superior performance. This shows our robust multi-tasking capabilities, demonstrating its effectiveness and potential as a world model.

\subsection{Real-World Deployment}
\begin{figure}[t]
    \centering
    \includegraphics[width=\columnwidth]{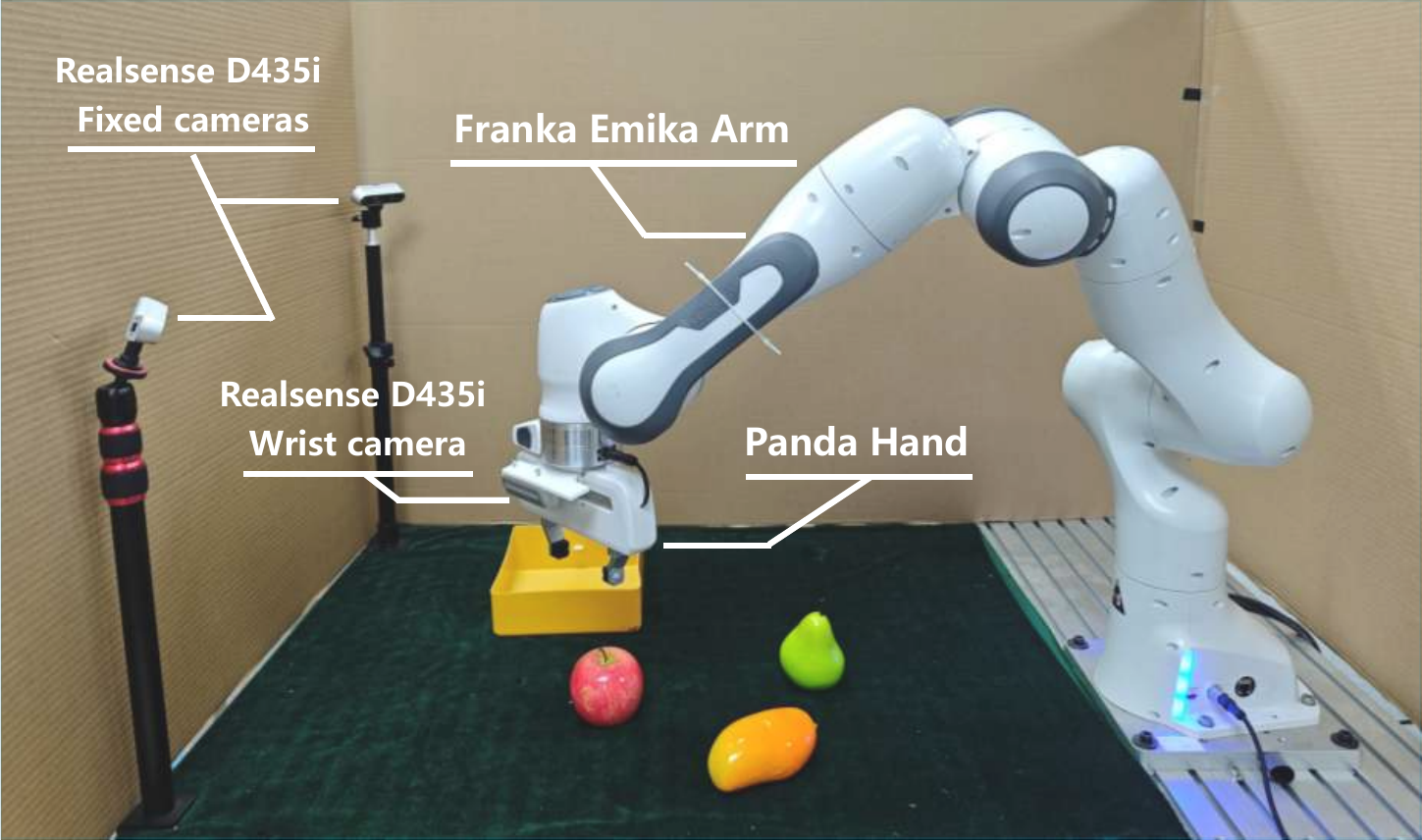}
    \caption{\textbf{Real-World Experiment Setup} comprising a Franka arm with Panda hand, equipped with two static cameras and one wrist-mounted camera for visual input.}
    \label{fig: realworld}
\end{figure}

\begin{figure*}[t]
    \centering
    \includegraphics[width=\textwidth]{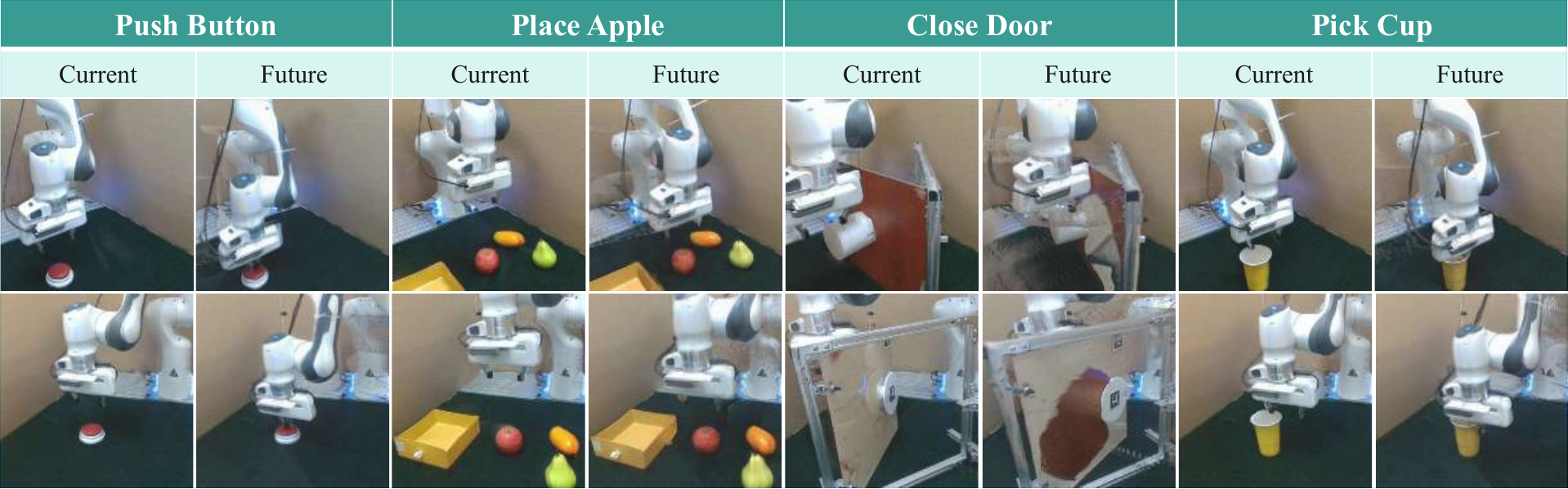}
    \caption{\textbf{Real-World Experiment GAF Results} : rendered images from current Gaussian and future Gaussian}
    \label{fig: realworld}
\end{figure*}


\textbf{Experiment Setup.} We evaluate our method using a real Franka Emika Panda robot equipped with Panda Hand on $5$ tasks as shown in Fig.~\ref{fig: realworld}. 
We use $3$ calibrated RealSense D435i cameras: two external camera and one mounted on the wrist of the robot. 
During data collection, we record camera intrinsics and extrinsics. GAF uses only intrinsics for 4D reconstruction, while the extrinsics are employed in the subsequent action refinement stage to render gripper poses for visual alignment.
For each task, we collect $20$ demonstrations which tries to guarantee a good coverage of the workspace.

\begin{wraptable}{r}{0.4\columnwidth}
\begin{tabular}{|l|l|}
\hline
Push   Button & 10 / 10 \\ \hline
Close   Door & 8 / 10 \\ \hline
Open   Door & 7 / 10 \\ \hline
Pick    Cup & 7 / 10 \\ \hline
Place   Apple & 6 / 10 \\ \hline
\end{tabular}%
\caption{\label{tab:rw} Success Rate in real world tasks.}
\end{wraptable}

\textbf{Results \& Discussion.} From Table~\ref{tab:rw}, we can see that our method is capable of completing tasks in the real world, where different noise sources such as imperfect camera calibration are present. 
Failure cases were primarily observed in the Place Apple in a Box and Open Door task.
Failures frequently occurred during the attempt to grasp the apple or door handle, likely due to the absence of force feedback. This limitation could potentially be addressed by augmenting the training data to better represent gripper state transitions or incorporating tactile sensing along with other sensors.

\subsection{Implement Details}
\textbf{Training Phase. }
GAF is end-to-end trained using RGB video frames for supervision. 
We initialize the ViT with the weights from MASt3R~\cite{leroy2024grounding}, while the remaining layers are initialized randomly. For a fair comparison, all methods, including the baselines, use observations (128 $\times$ 128 in simulator and 1280 $\times$ 720 in the real world) from two external cameras and another wrist camera. 
It have been trained for 80k iterations with a batch size of 16 using a single NVIDIA RTX A800 GPU, taking approximately 24 hours.

In the action denoising process, we use 50 diffusion ierations based on DDIM~\cite{song2020denoising}. To obtain more precise local observations, we incorporated the GT wrist camera data as an auxiliary resource. 
We use 1 last observations as input and predict 8 future actions. 
It have been trained for 50k iterations in 1.5 days on a single NVIDIA RTX A4090 GPU without extensive optimisation.

\textbf{Evaluation Phase. }
Different from training, during inference we only perform 3 diffusion iterations, making our online deployment more real-time. Each 8-step action prediction takes less than 0.3 seconds. In real-world deployment, we further eliminate inference-induced delays by running action prediction and execution in parallel using separate threads.

\section{Discussion}
We present a 4D representation GAF to formulate a V-4D-A paradigm that infers future scene evolution from current visual observations to guide robotic manipulation.
GAF supports scene reconstruction, future prediction, and action generation within a unified framework.
This feed-forward pipeline requires only sparse-view RGB images and supports real-time execution.
Experiments demonstrate that GAF achieves superior performance in both reconstruction  quality and success rate, and can be successfully deployed in real-world environments.
While our current method focuses on geometric modeling and motion prediction, it lacks semantic or task-level understanding.
Future work will incorporate language modeling to bring high-level semantic priors into the system to support context-aware manipulation.

\bibliographystyle{IEEEtran}
\bibliography{references}

\end{document}